\documentclass[10pt,twocolumn]{article}
\usepackage[margin=0.7in]{geometry}
\usepackage[hyphens]{url}  
\usepackage{graphicx} 
\usepackage{natbib}  
\usepackage{caption} 
\usepackage{algorithm}
\usepackage{algorithmic}

\usepackage{newfloat}
\usepackage{listings}
\DeclareCaptionStyle{ruled}{labelfont=normalfont,labelsep=colon,strut=off} 
\floatstyle{ruled}
\newfloat{listing}{tb}{lst}{}
\floatname{listing}{Listing}

\usepackage{booktabs}

\usepackage{times}
\usepackage{latexsym}
\usepackage{booktabs}
\usepackage[table]{xcolor}
\usepackage{array}
\usepackage{makecell}
\usepackage{booktabs}
\usepackage{multirow}
\usepackage{amsmath}
\usepackage{amssymb}
\usepackage{booktabs}
\usepackage{graphicx}
\usepackage{xcolor}
\usepackage{colortbl}
\usepackage{microtype}  
\usepackage{amsmath,amssymb}
\usepackage{booktabs,tabularx}
\usepackage{xcolor}
\usepackage{tikz,pgfplots}

\usepackage{xspace}
\usepackage{subcaption}
\usepackage{graphicx}
\newcommand{\model}{\textsc{TTIQ}\xspace}

\usepackage[T1]{fontenc}

\usepackage[utf8]{inputenc}

\usepackage{microtype}

\usepackage{inconsolata}

\usepackage{graphicx}

\title{Harnessing Image Question Dependence for Better VLM Test-time \\ Reinforcement Learning}

\author{
Xinrui He,
Ting-Wei Li,
Junting Wang,
Mengting Ai,
Xinyu He,
Hanghang Tong,
Jingrui He\\[4pt]
 University of Illinois Urbana-Champaign\\
 Champaign, IL, USA
}

\date{}

\begin{document}

\maketitle

\begin{abstract}
Test-time reinforcement learning can adapt vision-language models (VLMs) to unlabeled target data, but its effectiveness is fundamentally limited by the reliability of self-generated learning signals. To assess the reliability of consensus-based learning signals, we analyze VLM test-time reinforcement learning across diverse VQA datasets and model sizes, revealing two limitations. First, gains from consensus-based test-time training largely come from answer normalization rather than content correction. Second, many initial VLM responses are incorrect due to the model's limited ability to jointly use the image and the question; consensus rewards derived from these outputs may preserve the resulting grounding errors rather than correct them. Motivated by these, we propose \model{}, a test-time reinforcement learning framework that harnesses image-question dependence for better vlm adaptation. \model{} teacher-forces each sampled response under the original image-question pair and its image- and question-ablated variants, using the resulting token-level likelihood changes to estimate dependence on each input. It combines image and question dependence with calibrated confidence to
construct a response-level reward that favors jointly grounded responses, and uses the token-level signals to assign greater positive policy credit to tokens supported by both inputs. This design
favors responses that are jointly grounded in the image and the question and sufficiently confident, rather than merely popular. Experiments across eight VQA datasets and multiple VLM sizes show that \model{} achieves the best average performance at every model scale. It further generalizes across VLM families, while models trained on one dataset improve performance on unseen datasets without further training.
\end{abstract}


\section{Introduction}

Vision--language models (VLMs) have achieved remarkable progress on
multimodal understanding and reasoning tasks, including visual question answering over natural images, documents, and complex scenes \cite{goyal2017making, mathew2021docvqa}. Despite their
strong zero-shot capabilities, most VLMs remain static after deployment \cite{shu2022test}, as adapting to new domains or distribution shifts typically requires costly target-domain annotations and supervised fine-tuning. Test-time reinforcement learning offers a complementary approach by updating a pretrained generative model directly on unlabeled target inputs using learning signals available at inference time \citep{Wei2025FirstSS, zuo2026ttrl}. The central challenge is therefore how to construct a reliable learning signal without access to ground-truth answers.

\begin{figure}[t]
  \centering

  \begin{subfigure}[t]{0.48\columnwidth}
    \centering
    \includegraphics[width=\linewidth]{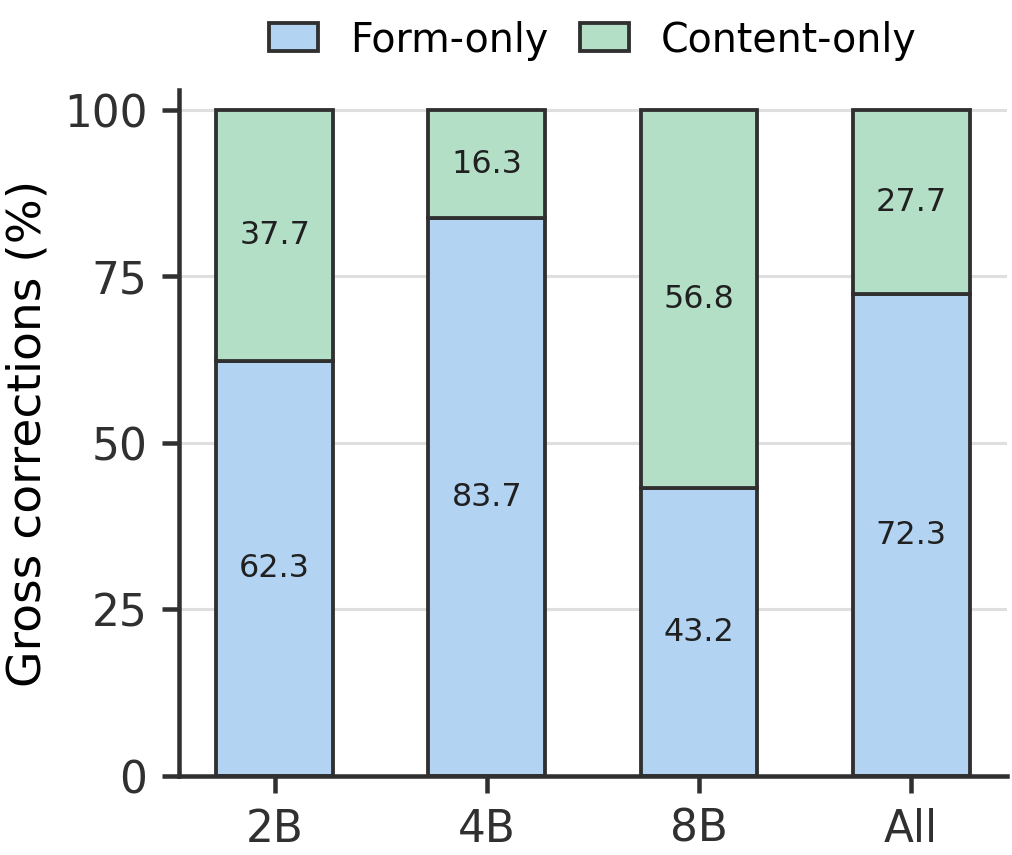}
    \caption{Composition of exact-match corrections.}
    \label{fig:ttrv_correction_type}
  \end{subfigure}
  \hfill
  \begin{subfigure}[t]{0.48\columnwidth}
    \centering
    \includegraphics[width=\linewidth]{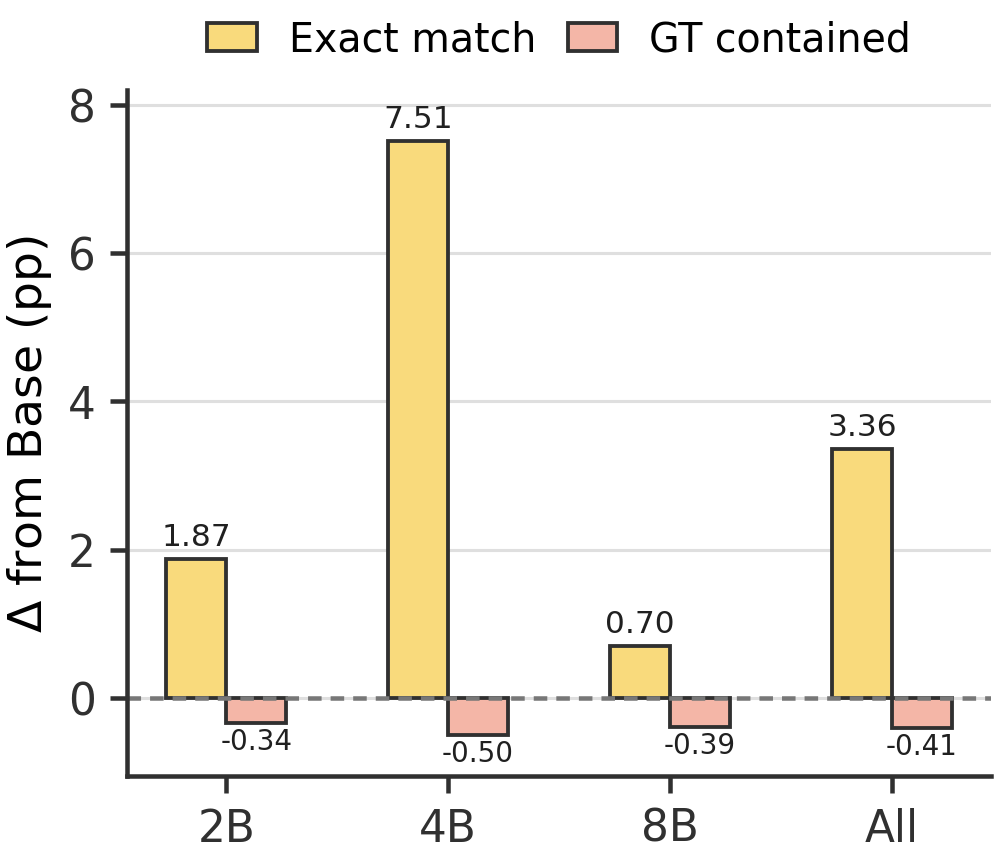}
    \caption{Changes in exact match and answer containment.}
    \label{fig:ttrv_metric_change}
  \end{subfigure}

\caption{
Analysis of where the gains of TTRV\citep{singh2026ttrv} come from across the eight-task evaluation.
In (a), an exact-match correction is considered form-only when the Base
response already contains an acceptable ground-truth span; 72.3\% of the
corrections fall into this category. In (b), TTRV improves exact match across
model scales, while ground-truth answer containment decreases. }
  \label{fig:ttrv_mechanism}
\end{figure}

\begin{figure}[t]
    \centering
    \includegraphics[width=0.98\linewidth]{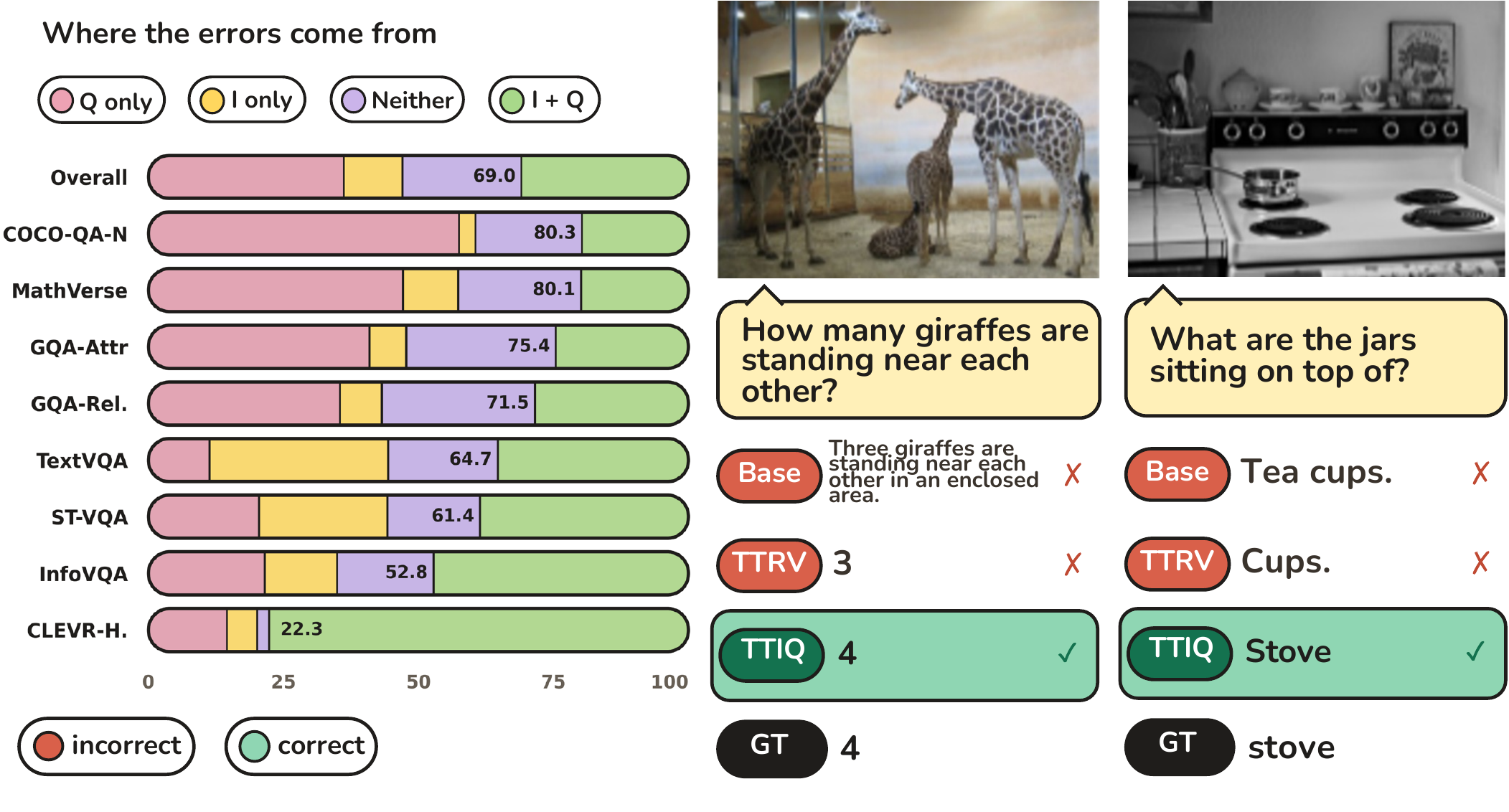}
    \caption{\textbf{Left:} Incorrect pre-update responses across eight tasks, grouped by dependence on the question only, the image only, neither input, or both inputs. Numbers indicate the percentage lacking joint image--question dependence. \textbf{Right:} Example predictions from Base, TTRV, and \model{}, together with the ground-truth answer.
    }
    \label{fig:motivation}
\end{figure}
Recent work has explored this challenge largely by deriving rewards from answer consensus across sampled outputs. Representative methods
include TTRL~\citep{zuo2026ttrl}, which treats the majority-vote answer as a pseudo-label for test-time reinforcement learning, while TTRV~\citep{singh2026ttrv} extends this principle to VLMs using response-frequency rewards and an entropy term over the empirical answer distribution. However, recent studies show that majority-vote supervision can amplify confirmation bias, favor already-solvable examples, and suppress minority correct signals \citep{wang2026beyond, lin2026detecting}. These concerns are especially relevant to open-ended VQA, where answers are generated freely. Since TTRV evaluates its design by standardizing tasks into a multiple-choice format, it remains unclear whether consensus-based test-time reinforcement learning in the more challenging free-form setting corrects answer content or mainly normalizes answer form. More fundamentally, although TTRV applies output-consensus rewards to VLMs, its reward remains modality-agnostic, relying on statistics over generated answers without explicitly incorporating multimodal information.

To characterize these gaps, we first conduct an in-depth analysis across
diverse datasets and VLM scales, summarized here and detailed in
Section~\ref{sec:motivation}. Our analysis reveals two recurring limitations. \textbf{First, the gains of consensus-based VLM test-time reinforcement learning largely come from answer normalization rather than content correction.} As shown in Figure~\ref{fig:ttrv_mechanism}, across model scales, most apparent corrections are form-only: exact-match accuracy improves even as ground-truth answer containment slightly decreases; 
\textbf{Second, most VLM errors exhibit insufficient joint use of the image and the question, which consensus rewards may then inherit.} As shown in Figure \ref{fig:motivation}, we examine the sampled responses used to construct consensus rewards and find that most incorrect responses lack dependence on at least one input. When such errors recur across rollouts, consensus may treat them as reliable signals and reinforce rather than correct them.

\begin{figure*}[ht]
    \centering
    \includegraphics[width=\linewidth]{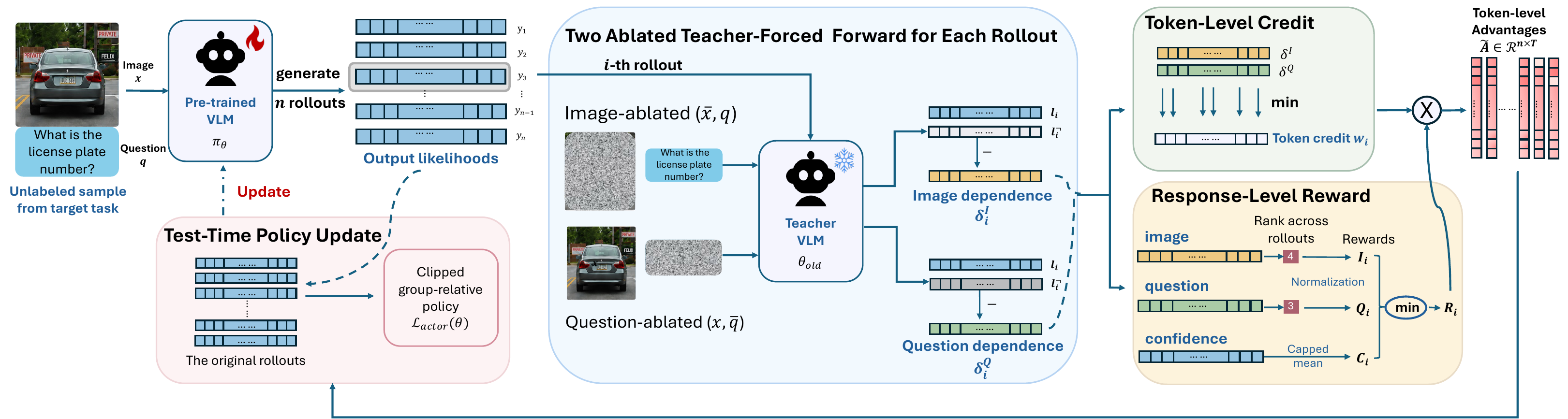}
    \caption{
Overview of \model{}. Given an unlabeled image--question pair, the VLM samples multiple responses. Each response is teacher-forced under image-ablated and question-ablated inputs to estimate token-level image and question dependence. These signals are used to construct a response-level reward and token-level credit assignment, followed by a test-time GRPO update.
}
    \label{fig:framework}
\end{figure*}

To address these challenges, we propose \model{}, a test-time reinforcement learning method for VLMs that rewards responses according
to their joint dependence on the image and the question, rather than agreement among sampled outputs. 
As illustrated in Figure~\ref{fig:framework}, \model{} teacher-forces each sampled response under the original input and its image- and question-ablated variants, and uses the resulting likelihood changes to estimate dependence on each input. At the \textbf{response level}, it combines these dependence signals with calibrated confidence to reward grounded and reliable responses. At the \textbf{token level}, it assigns greater positive policy credit to tokens supported by both inputs. This design shifts the learning signal from output
popularity toward responses that are jointly grounded in the image and the question and sufficiently confident.

Our contributions are:
\begin{itemize}
    \item We analyze consensus-based VLM test-time reinforcement learning across diverse VQA datasets and models, showing that its gains largely come from answer normalization rather than content correction, while many initial errors reflect limited joint use of the image and the question and can be preserved by consensus-based rewards.

    \item We introduce \model{}, the first test-time reinforcement learning method for generative VLMs to explicitly incorporate visual grounding into its training signal. It estimates image and question dependence through image and question ablations, and uses these signals for response-level reward construction and token-level credit assignment.

    \item Extensive experiments show \model{} achieves the best average performance at every evaluated model scale. It generalizes across VLM families, and models trained on one dataset consistently improve performance on unseen datasets without further training.
\end{itemize}

\section{Problem Setup}
\label{sec:problem_setup}

Let $x$ denote an image, $q$ a question, and
$\pi_\theta(y \mid x,q)$ an autoregressive vision--language model.
At test time, the pretrained model receives a small unlabeled support
set
\begin{equation}
\mathcal{U}
=
\{(x_j,q_j)\}_{j=1}^{J},
\end{equation}
drawn from the target task distribution. Ground-truth answers are not
available during adaptation.

At each training iteration, a frozen model $\pi_{\theta_{\mathrm{old}}}$ samples $N$ responses for each support
example:
\begin{align}
y_i
&=
(y_{i,1},\ldots,y_{i,T_i}),
\\
y_i
&\sim
\pi_{\theta_{\mathrm{old}}}
(\cdot \mid x,q),
\qquad i=1,\ldots,N.
\end{align}

The goal is to update $\pi_\theta$ through reinforcement learning using rewards constructed solely from the unlabeled support set and sampled responses, thereby improving performance on the target task.
\section{Method}
\label{sec:method}

In this section, we first analyze consensus-based VLM test-time training to motivate joint image--question dependence (Section~\ref{sec:motivation}). We then present our method, including the response-level reward
(Section~\ref{sec:response_reward}), joint-token credit assignment (Section~\ref{sec:token_credit}), and GRPO-based update (Section~\ref{sec:optimization}).

\subsection{Limitations of Output-Consensus Signal}
\label{sec:motivation}
\paragraph{Consensus gains largely come from answer normalization.}
We first examine what consensus-based VLM test-time training changes in model behavior and performance.  Across the 24 full-test model--dataset settings in our main evaluation, covering eight VQA datasets and three VLM scales, we define an exact-match correction as a prediction for which Base is exact-match incorrect but TTRV is exact-match correct. We identify 4,131 such corrections in total. As shown in Figure~\ref{fig:ttrv_mechanism}, 72.3\% are form-only corrections, meaning that the Base response already contains a ground-truth answer span, while TTRV mainly adjusts the answer boundary or canonical form to satisfy exact matching. Consistent with this finding, TTRV improves macro exact-match accuracy by 3.36 points, whereas ground-truth answer containment decreases by 0.41 points. Figure~\ref{fig:motivation} (right) provides representative examples. TTRV converts ``A car'' to ``Car,'' compresses an incorrect count of three giraffes into ``3,'' and simplifies ``Tea cups'' to ``Cups.'' These results show that much of the measured improvement comes from answer normalization rather than new content correction. The responses become easier to match without becoming more semantically correct. More examples are provided in Appendix D.

\paragraph{Consensus is constrained by VLM's initial failures.}

We next examine a more fundamental limitation of output-consensus supervision. Because consensus rewards are derived from the model's own sampled responses, their effectiveness is limited by the quality of those responses. When the initial responses are incorrect, agreement among them may preserve the error rather than improve the model's underlying capability. This concern is particularly relevant for VLM errors, as prior work shows that when generating responses, models tends to rely heavily on language priors while underusing visual evidence \citep{asadi2026mirage,lee2025vlind}. We conduct a diagnostic with InternVL2.5-4B using 200 examples from each of eight datasets and four rollouts per example, yielding 6,400 responses. For each response, we keep its tokens fixed and compare their likelihoods under the original input, a zero-image input, and a question-masked input. A sustained likelihood drop after ablating an input indicates dependence on that input. As shown in Figure~\ref{fig:motivation}, 69.0\% of incorrect responses lack dependence on at least one input. This finding identifies limited joint use of the image and the question as one of the major sources of VLM errors.

\subsection{Response Reward from Image--Question Dependence}
\label{sec:response_reward}

Based on the findings, we propose a training reward based on joint image--question dependence. Given a sampled response $y_i=(y_{i,1},\ldots,y_{i,T_i})$, where $i$ indexes responses and $t$ indexes tokens, we score the same fixed token sequence under three inputs: the original image--question pair, a zero-image input $\bar{x}$, and a question-masked input $\bar{q}$. To construct $\bar{x}$, we replace the pixel tensor with an all-zero tensor of the same shape, while retaining the visual-token layout. To construct $\bar{q}$, we replace only the question content tokens with the tokenizer's padding token, while preserving the sequence length. We hold the response fixed to isolate how much each input supports that response. The corresponding token log-likelihoods are
\begin{equation}
\begin{aligned}
\ell_{i,t}
&=
\log \pi_{\theta_{\mathrm{old}}}
\left(y_{i,t}\mid x,q,y_{i,<t}\right),\\
\ell^{\neg I}_{i,t}
&=
\log \pi_{\theta_{\mathrm{old}}}
\left(y_{i,t}\mid \bar{x},q,y_{i,<t}\right),\\
\ell^{\neg Q}_{i,t}
&=
\log \pi_{\theta_{\mathrm{old}}}
\left(y_{i,t}\mid x,\bar{q},y_{i,<t}\right).
\end{aligned}
\label{eq:three_likelihoods}
\end{equation}

We define token-level image and question dependence as
\begin{equation}
\delta^I_{i,t}
=
\ell_{i,t}-\ell^{\neg I}_{i,t},
\qquad
\delta^Q_{i,t}
=
\ell_{i,t}-\ell^{\neg Q}_{i,t}.
\label{eq:token_dependence}
\end{equation}
Because the response is fixed, each ablated condition requires only one teacher-forced forward pass, without additional autoregressive decoding. A positive likelihood difference indicates that the corresponding input increases the likelihood assigned to that response token.

\paragraph{Aggregating token dependence.}
A useful response should depend on an input across its entire answer, rather than through only a few isolated tokens. For the $i$-th response of length $T_i$ and modality $m\in\{I,Q\}$, let
\[
k_i=\left\lceil\frac{T_i}{2}\right\rceil,
\qquad
\mathcal{L}_i^m
=
\operatorname{Low}_{k_i}({\delta}_i^m),
\]
where $\operatorname{Low}_{k}(\cdot)$ returns the indices of the $k$ smallest finite values. We aggregate this token-level vector into two response-level statistics that capture the response's overall dependence and sustained dependence:
\begin{equation}
\begin{aligned}
d_i^m
=
\frac{1}{T_i}
\sum_{t=1}^{T_i}\delta^m_{i,t},\quad
s_i^m
=
\frac{1}{k_i}
\sum_{t\in\mathcal{L}_i^m}\delta^m_{i,t}.
\end{aligned}
\label{eq:dependence_aggregation}
\end{equation}
The mean $d_i^m$ captures overall dependence level, while the lower-half average $s_i^m$ measures whether that dependence is sustained across the response rather than concentrated in a few tokens.

Raw likelihood changes can vary in scale with response content and model confidence. We therefore compare responses using within-group ranks. For a rollout group
$\mathcal{G}$ containing $n$ valid responses, let $\rho_{\mathcal{G}}(z_i)\in[0,n-1]$ denote the zero-based average rank of $z_i$. We define
\begin{equation}
\mathcal{R}_{\mathcal{G}}(z_i)
=
2\frac{\rho_{\mathcal{G}}(z_i)}{n-1}-1.
\label{eq:rank_normalization}
\end{equation}
This transformation maps the lowest-ranked response to $-1$ and the highest-ranked response to $1$, while preserving the relative ordering of responses. We next apply this rank transformation to the response-level dependence scores.

The response-level image and question dependence scores are then
\begin{equation}
\begin{aligned}
I_i
&=
\mathcal{R}_{\mathcal{G}}
\left(
\mathcal{R}_{\mathcal{G}}(d_i^I)
+
\mathcal{R}_{\mathcal{G}}(s_i^I)
\right),\\
Q_i
&=
\mathcal{R}_{\mathcal{G}}
\left(
\mathcal{R}_{\mathcal{G}}(d_i^Q)
+
\mathcal{R}_{\mathcal{G}}(s_i^Q)
\right).
\end{aligned}
\label{eq:response_dependence}
\end{equation}

\paragraph{Minimum-bottleneck reward.}
Image and question dependence scores measure how each input supports a response. We additionally measure the model's confidence score in the response using its length-normalized likelihood under the original input:
\begin{equation}
C_i
=
\mathcal{R}_{\mathcal{G}}
\left[
\exp\left(
\frac{1}{T_i}
\sum_{t=1}^{T_i}
\ell_{i,t}
\right)
\right].
\label{eq:confidence}
\end{equation}
The exponential converts the mean token log-likelihood into a length-normalized sequence probability, with a higher $C_i$ indicating stronger model confidence for the complete response. We incorporate confidence score through
\begin{equation}
C^{\mathrm{cap}}_{i,\lambda}
=
(1-\lambda)+\lambda C_i,
\qquad
\lambda\in[0,1],
\label{eq:confidence_cap}
\end{equation}
Here, $\lambda$ controls how strongly confidence score is allowed to limit the reward. When $\lambda=0$, the cap is fixed at $1$ and confidence has no effect; increasing $\lambda$ allows a low-confidence response to impose a stronger upper bound. The final response level reward is
\begin{equation}
R_i
=
\mathcal{R}_{\mathcal{G}}
\left(
\min
\left\{
I_i,\,
Q_i,\,
C^{\mathrm{cap}}_{i,\lambda}
\right\}
\right).
\label{eq:response_reward}
\end{equation}
The minimum forms a bottleneck: a response receives a high reward only when it depends strongly on both the image and the question and is sufficiently confident. High confidence cannot independently increase the reward, whereas low confidence can constrain it.

\subsection{Joint-Token Credit Assignment}
\label{sec:token_credit}

The above response-level reward determines whether a sampled response should be reinforced or suppressed, but it does not distinguish the contributions of individual tokens. A positively rewarded response may contain grounded answer tokens together with generic templates, formatting phrases, or unsupported explanations. Applying the same positive advantage to every token would reinforce all of them equally. We therefore redistribute positive credit according to token level image--question dependence. For negatively rewarded responses, however, we retain the full response-level penalty for every token so that no part of a disfavored response is exempted from suppression.

We first standardize the response rewards using the mean and standard deviation within each rollout group to obtain the response-level advantage $A_i=(R_i-\mu_R)/(\sigma_R+\epsilon)$. For each token, let $P_i^I(t)$ and $P_i^Q(t)$ denote the percentile ranks of its image and question dependence scores, respectively. We define its joint support as
\begin{equation}
g_{i,t}
=
\min
\left\{
P_i^I(t), P_i^Q(t)
\right\}.
\label{eq:token_support}
\end{equation}
We normalize $g_{i,t}$ into a positive credit weight $w_{i,t}$ whose mean within each response equals one. This changes how positive credit is distributed across tokens without changing the total positive credit assigned to the response.

The resulting token-level joint-token advantage is 
\begin{equation}
\widetilde{A}_{i,t}
=
\begin{cases}
w_{i,t}A_i, & A_i>0,\\
A_i, & A_i\leq 0.
\end{cases}
\label{eq:token_advantage}
\end{equation}
For preferred responses, tokens with stronger joint support receive more positive credit. For a disfavored response, every token retains the same negative response-level advantage and is suppressed uniformly.

\begin{table*}[t]
  \centering
  \resizebox{\textwidth}{!}{%
  \begin{tabular}{@{}lrrrrrrrrr@{}}
    \toprule
    Method & InfoVQA & TextVQA & GQA-Rel. & GQA-Attr & ST-VQA & MathVerse-VD & COCO-QA-N & CLEVR-H & Mean \\
    \midrule
    \rowcolor{black!7} LLaMA-3.2-11B & 46.36 & 62.59 & 2.76 & 14.72 & 61.55 & 8.72 & 36.38 & 24.19 & 32.16 \\
    \rowcolor{black!7} LLaVA-1.5-7B & 15.48 & 53.43 & 41.54 & 56.30 & 43.06 & 14.58 & 63.80 & 36.33 & 40.57 \\
    \rowcolor{black!7} Qwen2.5-VL-3B & 66.06 & 81.87 & 37.41 & 55.61 & 71.25 & 28.65 & 73.75 & 80.60 & 61.90 \\
    \rowcolor{black!7} Qwen2.5-VL-7B & 70.62 & 84.40 & 33.69 & 47.42 & 76.69 & 32.16 & 78.79 & 80.70 & 63.06 \\
    \rowcolor{black!7} Qwen3-VL-2B & 62.30 & 80.52 & 34.21 & 48.91 & 69.17 & 27.08 & 72.80 & 71.69 & 58.34 \\
    \rowcolor{black!7} Qwen3-VL-4B & 70.07 & 81.83 & 41.00 & 54.93 & 71.56 & 30.73 & 74.99 & 79.98 & 63.14 \\
    \midrule
    InternVL3-2B & 51.90 & 76.27 & 26.08 & 46.91 & 75.55 & 25.91 & 67.53 & 65.40 & 54.44 \\
    w/ TTRL & \textbf{53.30} & \underline{78.71} & 33.20 & \underline{51.72} & \underline{76.32} & \underline{26.95} & 72.72 & 66.92 & \underline{57.48} \\
    w/ SCOPE & \underline{53.21} & 78.45 & \underline{33.67} & 51.60 & 76.29 & 26.69 & 72.54 & \underline{67.08} & 57.44 \\
    w/ TTRV & 52.63 & 78.37 & 25.66 & \underline{51.72} & 76.16 & 26.43 & \underline{73.89} & 65.62 & 56.31 \\
    \rowcolor{blue!6} \textbf{w/ \model{}~(Ours)} & \underline{53.21} & \textbf{79.48} & \textbf{35.11} & \textbf{53.09} & \textbf{77.69} & \textbf{27.21} & \textbf{75.14} & \textbf{69.19} & \textbf{58.77} \\
    \midrule
    InternVL2.5-4B & 57.89 & 73.55 & 17.25 & 45.59 & 80.07 & 28.91 & 36.01 & 78.74 & 52.25 \\
    w/ TTRL & \textbf{59.79} & \underline{79.32} & \underline{35.71} & 56.24 & \underline{82.00} & \underline{29.43} & 68.78 & \underline{78.84} & 61.26 \\
    w/ SCOPE & 59.67 & 78.82 & 35.35 & \underline{56.99} & 81.98 & 28.91 & \underline{70.13} & \textbf{79.06} & \underline{61.36} \\
    w/ TTRV & 58.35 & 74.76 & 31.06 & 55.50 & 81.84 & 29.17 & 69.58 & 77.85 & 59.76 \\
    \rowcolor{blue!6} \textbf{w/ \model{}~(Ours)} & \underline{59.73} & \textbf{79.52} & \textbf{37.15} & \textbf{57.10} & \textbf{82.67} & \textbf{29.56} & \textbf{74.70} & 78.81 & \textbf{62.41} \\
    \midrule
    InternVL3-8B & 65.58 & 84.10 & 35.84 & 57.22 & 84.49 & 33.46 & 80.51 & 84.17 & 65.67 \\
    w/ TTRL & 65.42 & \underline{84.54} & 39.58 & \underline{59.56} & \underline{84.72} & 34.38 & 80.48 & \underline{84.39} & 66.63 \\
    w/ SCOPE & \textbf{65.82} & \textbf{84.90} & \underline{39.79} & 59.11 & 84.71 & 34.11 & \underline{80.55} & 84.36 & \underline{66.67} \\
    w/ TTRV & 65.51 & 83.86 & 39.66 & 58.13 & 84.52 & \underline{34.51} & 80.37 & \underline{84.39} & 66.37 \\
    \rowcolor{blue!6} \textbf{w/ \model{}~(Ours)} & \underline{65.76} & 84.48 & \textbf{40.10} & \textbf{60.31} & \textbf{84.84} & \textbf{34.77} & \textbf{80.80} & \textbf{84.50} & \textbf{66.94} \\
    \bottomrule
  \end{tabular}%
  }
  \caption{Comparison with all test-time reinforcement learning baselines. Best and second-best results within each backbone block are shown in bold and underlined.}
  \label{tab:main_full_vqa_all}
\end{table*}

\subsection{Test-Time Policy Update}
\label{sec:optimization}

We optimize the model with the standard clipped GRPO objective \citep{shao2024deepseekmath}, replacing the uniform response-level advantage with $\widetilde{A}_{i,t}$ from Eq.~\eqref{eq:token_advantage}. For each generated token, the policy ratio under the original input is
\begin{equation}
r_{i,t}(\theta)
=
\frac{
\pi_{\theta}
\left(
y_{i,t}\mid x,q,y_{i,<t}
\right)
}{
\pi_{\theta_{\mathrm{old}}}
\left(
y_{i,t}\mid x,q,y_{i,<t}
\right)
}.
\label{eq:policy_ratio}
\end{equation}
The training objective is
\begin{equation}
\begin{aligned}
\mathcal{L}(\theta)
={}&
-\mathbb{E}_{i,t}
\left[
\min
\left\{
\begin{aligned}
&r_{i,t}(\theta)\widetilde{A}_{i,t},\\
&\operatorname{clip}\!\left(
r_{i,t}(\theta),
1-\varepsilon,
1+\varepsilon
\right)
\widetilde{A}_{i,t}
\end{aligned}
\right\}
\right]
\\
&+\beta\mathcal{L}_{\mathrm{KL}} .
\end{aligned}
\label{eq:grpo_objective}
\end{equation}
The policy ratio is always evaluated using the original input; the ablated inputs are used only to construct the
response reward and token-level credit assignment.

\paragraph{Training efficiency.}
Relative to standard GRPO, \model{} adds only two teacher-forced scoring passes, one for each controlled input ablation. Because the response sequence is fixed, all token likelihoods are computed in parallel without additional autoregressive decoding.  Appendix~B provides detailed training-time comparisons and analyzes how the number of training examples affects performance and runtime.

\section{Experimental Results}

\subsection{Experiment Setup}

\paragraph{Datasets and Baselines.}

We evaluate TTIQ on eight VQA tasks: InfoVQA \cite{mathew2022infographicvqa}, TextVQA \cite{singh2019towards}, ST-VQA \cite{biten2019scene}, the two largest VQA subsets in GQA \cite{hudson2019gqa}, Relation and Attribute, MathVerse Vision Dominant \cite{zhang2024mathverse}, COCO-QA Number \cite{ren2015exploring}, and CLEVR-Humans \cite{johnson2017clevr,johnson2017inferring}. Our primary baselines are Base model, which performs deterministic zero-shot inference with the base VLM model, and TTRV \cite{singh2026ttrv}, the closest prior test-time reinforcement-learning method for VLMs. We additionally compare with two consensus-based methods originally developed for LLM test-time reinforcement learning and adapted here to VLM, TTRL~\citep{zuo2026ttrl}, which reinforces the majority-vote answer as a pseudo-label, and SCOPE~\citep{wang2026beyond}, which weighting the consensus signal with the model's own confidence.  We also report zero-shot Llama-3.2-11B-Vision \cite{grattafiori2024llama}, LLaVA-1.5-7B \cite{liu2024improved}, Qwen2.5-VL-3B/7B \cite{Bai2025Qwen25VLTR}, and Qwen3-VL-2B/4B \cite{bai2025qwen3} as VLM reference points. For fair comparisons, all methods use the same prompts, answer extraction, procedure and shared hyperparameters. For all experiments, we report exact match accuracy (\%), We also report token-F1 and ROUGE-L results, together with a detailed
decomposition of \model{}'s improvements, in Appendix~C.

\paragraph{Implementation.}
We implement \model{} with InternVL3-2B, InternVL2.5-4B, and InternVL3-8B \cite{chen2024expanding, zhu2025internvl3}, and evaluate cross-family transfer on Qwen2.5-VL-3B/7B. Following TTRV, for each task, we randomly sample 20 examples as unlabeled training data, and evaluate each remaining example in the corresponding evaluation split. Following TTRL and TTRV, we sample 32 responses per prompt at temperature 1.0, retain 16 for training, and use an actor learning rate of $5\times10^{-7}$ with cosine decay and 3\% warmup; we reuse the prompt provided by TTRL and TTRV, and prompt and response lengths are capped at 7,524 and 1,024 tokens. For joint-token credit, we map each token's joint support to $w'_{i,t}=0.2+0.8g_{i,t}$ and then normalize the weights, preserving nonzero credit for weakly grounded tokens. Evaluation uses greedy decoding. For Table \ref{tab:main_full_vqa_all}, we report the best observed $\lambda$ over $\lambda\in {0,0.25,0.5,0.75,1}$ which is our only tuned hyperparameter. All runs use NVIDIA A100 and GH200 GPUs.

\subsection{Main Results}

Table~\ref{tab:main_full_vqa_all} shows that TTIQ achieves the highest average accuracy at every model scale. More importantly, TTIQ outperforms TTRV in all 24 model--dataset settings. From these results, we make three observations.

First, TTIQ yields particularly large gains on tasks requiring fine-grained image--question grounding, including visual relations, attributes, and counting. Averaged over the nine GQA-Relation, GQA-Attribute, and COCO-QA Number settings, TTIQ improves over Base by 11.18 points and over TTRV by 3.10 points. Case-level comparisons show that TTIQ corrects object identities, attributes, and counts even when Base and TTRV give the same wrong answer. This suggests that the gains go beyond answer reformatting and are consistent with better joint use of the image and  question. Second, the benefit of test-time adaptation is smaller on the strongest 8B backbone than on the 2B and 4B models. This suggests that stronger pretrained models already possess better multimodal understanding, leaving less room for further improvement at test time. Nevertheless, TTIQ improves all eight tasks with the 8B backbone, showing that it remains effective even when the initial model is strong. Third, TTIQ remains effective across different task types. Beyond its larger gains on fine-grained image--question grounding, it outperforms TTRV in all
nine OCR-heavy settings and all three MathVerse Vision Dominant settings, while improving or matching Base throughout. This shows that emphasizing joint image--question dependence generalizes across text recognition and visual mathematical reasoning.

\subsection{Generalization}

\begin{table}[t]
  \centering
  \begingroup
  \setlength{\tabcolsep}{2.5pt}
  \renewcommand{\arraystretch}{1.02}

  \resizebox{\columnwidth}{!}{%
  \begin{tabular}{@{}llrrrr@{}}
    \toprule
    Backbone & Method
      & \multicolumn{3}{c}{Target dataset}
      & Avg. \\
    \midrule

    \multicolumn{2}{l}{\textbf{Source: GQA-Attr}}
      & \textbf{GQA-Rel}
      & \textbf{COCO-N}
      & \textbf{MathV-VD}
      & \\

    IVL3-2B
      & Base
      & 26.08 & 67.53 & 25.91 & 39.84 \\
    IVL3-2B
      & TTRV
      & 29.46 & 72.72 & 26.82 & 43.00 \\
    \rowcolor{blue!6}
    IVL3-2B
      & \textbf{TTIQ}
      & \textbf{30.34}
      & \textbf{73.31}
      & \textbf{26.95}
      & \textbf{43.53} \\

    IVL2.5-4B
      & Base
      & 17.25 & 36.01 & 28.91 & 27.39 \\
    IVL2.5-4B
      & TTRV
      & 30.00 & 71.12 & 29.17 & 43.43 \\
    \rowcolor{blue!6}
    IVL2.5-4B
      & \textbf{TTIQ}
      & \textbf{33.44}
      & \textbf{74.19}
      & \textbf{29.43}
      & \textbf{45.68} \\

    \midrule

    \multicolumn{2}{l}{\textbf{Source: GQA-Rel}}
      & \textbf{GQA-Attr}
      & \textbf{COCO-N}
      & \textbf{MathV-VD}
      & \\

    IVL3-2B
      & Base
      & 46.91 & 67.53 & 25.91 & 46.78 \\
    IVL3-2B
      & TTRV
      & 50.46 & 71.66 & 26.17 & 49.43 \\
    \rowcolor{blue!6}
    IVL3-2B
      & \textbf{TTIQ}
      & \textbf{51.55}
      & \textbf{74.37}
      & \textbf{26.43}
      & \textbf{50.78} \\

    IVL2.5-4B
      & Base
      & 45.59 & 36.01 & 28.91 & 36.84 \\
    IVL2.5-4B
      & TTRV
      & 54.41 & 66.54 & 29.56 & 50.17 \\
    \rowcolor{blue!6}
    IVL2.5-4B
      & \textbf{TTIQ}
      & \textbf{56.70}
      & \textbf{75.28}
      & \textbf{30.21}
      & \textbf{54.06} \\

    \midrule

    \multicolumn{2}{l}{\textbf{Source: TextVQA}}
      & \textbf{GQA-Rel}
      & \textbf{GQA-Attr}
      & \textbf{COCO-N}
      & \\

    IVL3-2B
      & Base
      & 26.08 & 46.91 & 67.53 & 46.84 \\
    IVL3-2B
      & TTRV
      & 29.72 & 48.51 & 72.91 & 50.38 \\
    \rowcolor{blue!6}
    IVL3-2B
      & \textbf{TTIQ}
      & \textbf{30.52}
      & \textbf{49.14}
      & \textbf{73.67}
      & \textbf{51.11} \\

    IVL2.5-4B
      & Base
      & 17.25 & 45.59 & 36.01 & 32.95 \\
    IVL2.5-4B
      & TTRV
      & 16.91 & 48.91 & 33.53 & 33.12 \\
    \rowcolor{blue!6}
    IVL2.5-4B
      & \textbf{TTIQ}
      & \textbf{30.16}
      & \textbf{55.67}
      & \textbf{72.39}
      & \textbf{52.74} \\

    \bottomrule
  \end{tabular}%
  }

  \endgroup
  \caption{Cross-dataset generalization of TTIQ. Bold values indicate the best result for each setting.}
  \label{tab:iqc_cross_generalization}
\end{table}

\paragraph{Across datasets.}
Our method also generalizes well to unseen datasets after being trained on a single source dataset. We train InternVL3-2B and InternVL2.5-4B on one source dataset and evaluate them directly on unseen target datasets. As shown in Table~\ref{tab:iqc_cross_generalization}, TTIQ improves over the base model and TTRV on all target datasets. The improvement also transfers between substantially different tasks. For example, adapting the 4B model on GQA-Relation improves its accuracy from 36.01 to 75.28 on COCO-QA-N and from 28.91 to 30.21 on MathVerse-VD. These results suggest that the learned updates strengthen the model's general ability, rather than merely fitting the answer distribution of the source dataset.
\paragraph{Across VLM families.}
We further evaluate \model{} on two model sizes from another VLM family, Qwen2.5-VL-3B and Qwen2.5-VL-7B. As shown in Table~\ref{tab:qwen25_transfer_full}, \model{} achieves the best average performance at both scales and is best or nearly tied on every task. These results show that \model{} generalizes beyond InternVL to a different VLM family and remains effective across model sizes.

\begin{table}[t]
  \centering
  
  \begingroup

  \setlength{\tabcolsep}{3pt}
  \begin{tabular}{@{}lrrrrr@{}}
    \toprule
    Method & InfoVQA & GQA-Attr & GQA-Rel & COCO-N & Avg. \\
    $n$    & 3,268   & 1,746    & 3,873   & 2,735  &      \\
    \midrule

    \multicolumn{6}{c}{\textbf{Qwen2.5-VL-3B}} \\
    Base
      & \textbf{66.06} & 55.61 & 37.41 & 73.75 & 58.21 \\
    TTRV
      & 65.97 & 55.44 & 37.31 & 73.49 & 58.05 \\
    \rowcolor{blue!6}
    \textbf{TTIQ}
      & 66.03 & \textbf{55.90} & \textbf{37.70}
      & \textbf{74.19} & \textbf{58.45} \\

    \midrule
    \multicolumn{6}{c}{\textbf{Qwen2.5-VL-7B}} \\
    Base
      & 70.62 & 47.42 & 33.69 & 78.79 & 57.63 \\
    TTRV
      & 70.53 & 50.11 & 36.97 & 78.90 & 59.13 \\
    \rowcolor{blue!6}
    \textbf{TTIQ}
      & \textbf{70.87} & \textbf{51.78} & \textbf{37.67}
      & \textbf{79.01} & \textbf{59.83} \\

    \bottomrule
  \end{tabular}
  \endgroup
  \caption{
  Generalization of \model{} to Qwen2.5-VL.
  }
  \label{tab:qwen25_transfer_full}
\end{table}

\begin{table*}[t]
  \centering
  \begingroup
  \small
  \setlength{\tabcolsep}{4.5pt}
  \begin{tabular}{@{}lllrrrrr@{}}
    \toprule
    Method & Reward & Credit
      & InfoVQA & TextVQA & GQA-Attr & GQA-Rel. & Avg. \\
    \midrule
    \multicolumn{8}{@{}l}{\textit{Reference methods}} \\
    Base & \multicolumn{1}{c}{} & \multicolumn{1}{c}{}
      & 57.89 & 73.55 & 45.59 & 17.25 & 48.57 \\
    TTRV & answer frequency & \multicolumn{1}{c}{}
      & 58.35 & 74.76 & 55.50 & 31.06 & 54.92 \\
    \midrule
    \multicolumn{8}{@{}l}{\textit{Reward components}} \\
    Image only & \(I\) & full response
      & 59.49 & 78.71 & 56.13 & 36.33 & 57.66 \\
    Question only & \(Q\) & full response
      & 57.65 & 78.41 & 56.93 & 23.96 & 54.24 \\
    w/o image & \(\min(Q,C)\) & full response
      & 58.57 & 73.96 & 56.47 & 15.90 & 51.23 \\
    w/o question & \(\min(I,C)\) & full response
      & 58.14 & 73.43 & 44.04 & 16.24 & 47.96 \\
    Mean fusion & \(\operatorname{mean}(I,Q,C)\) & full response
      & 57.77 & 73.67 & 44.73 & 17.07 & 48.31 \\
    w/o sustained dependence & \(\min(I_{\mathrm{overall}},Q_{\mathrm{overall}},C)\)
      & full response & 57.62 & 73.43 & 45.19 & 17.04 & 48.32 \\
    \midrule
    \multicolumn{8}{@{}l}{\textit{Credit assignment; full reward fixed}} \\
    Full-response credit & \(\min(I,Q,C)\) & full response
      & 59.49 & 76.37 & 56.19 & 33.57 & 56.40 \\
    Joint-token credit & \(\min(I,Q,C)\) & joint token
      & 59.61 & 76.51 & 56.99 & 35.48 & 57.14 \\
    \midrule
    \multicolumn{8}{@{}l}{\textit{Confidence component}} \\
    w/o confidence & \(\min(I,Q)\) & joint token
      & 59.64 & 77.95 & 57.04 & 36.53 & 57.79 \\
    \midrule
    \rowcolor{blue!6}
    \textbf{TTIQ} & \(\min(I,Q,C_{\lambda>0})\) & joint token
      & \textbf{59.73} & \textbf{79.52} & \textbf{57.10}
      & \textbf{37.15} & \textbf{58.38} \\
    \bottomrule
  \end{tabular}
  \endgroup

  \caption{Component ablation on InternVL2.5-4B. Bold indicates the best result in each column over the entire table.}
    \vspace{-10pt}
  \label{tab:ablation}
\end{table*}
\subsection{Ablation Study}
We conduct component ablations on InternVL2.5-4B over four representative tasks. All variants use the same training split and common hyperparameters and use eight updates. First, we vary the reward construction, then compare full-response credit with joint-token credit. Finally, we study the contribution of confidence. 

Table~\ref{tab:ablation} yields three main observations. First, image and question dependence both contribute to robust reward construction. Removing either signal reduces average accuracy. Although the image-only reward remains competitive overall, the question-only reward degrades substantially on GQA-Relation, showing that single-input signals are not consistently reliable across tasks. Second, the aggregation design is important. Minimum fusion outperforms mean fusion, while incorporating sustained dependence improves over using only the overall dependence level by 8.08 points. These results support requiring every reward component to be strong and measuring whether dependence persists across the response. Third, joint-token credit and calibrated confidence provide complementary improvements. Token-level credit better reinforces tokens supported by both inputs, while confidence further filters insufficiently reliable responses. Combining all components gives the strongest overall performance. 

\subsection{Sensitivity Analysis}
\label{sec:confidence_sensitivity}

\begin{figure}[t]
  \centering
  \captionsetup[subfigure]{skip=0pt}

  \begin{subfigure}[t]{0.475\columnwidth}
    \centering
    \includegraphics[width=\linewidth]
    {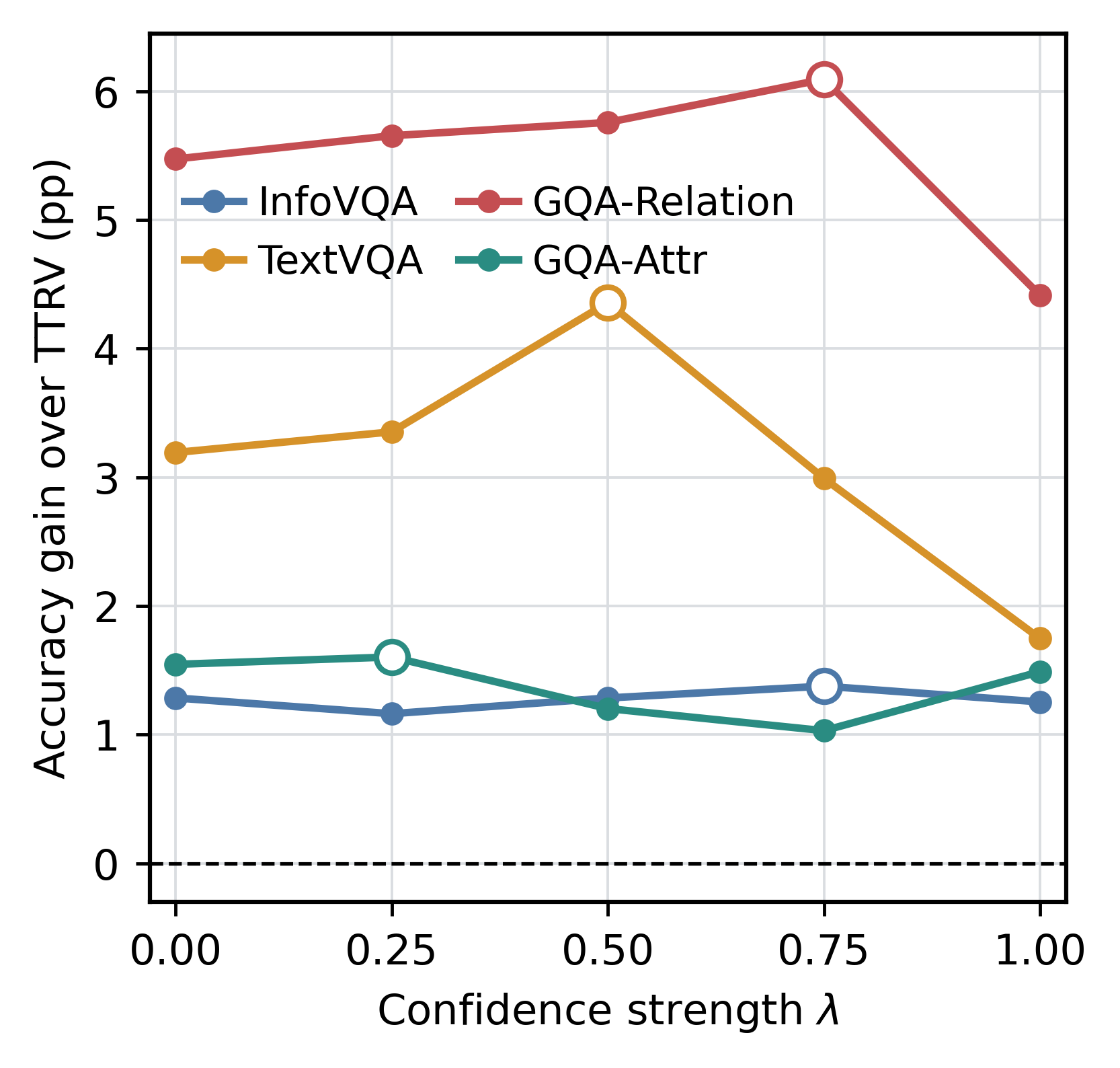}
    \caption{}
  \end{subfigure}
  \hfill
  \begin{subfigure}[t]{0.49\columnwidth}
    \centering
    \includegraphics[width=\linewidth]
    {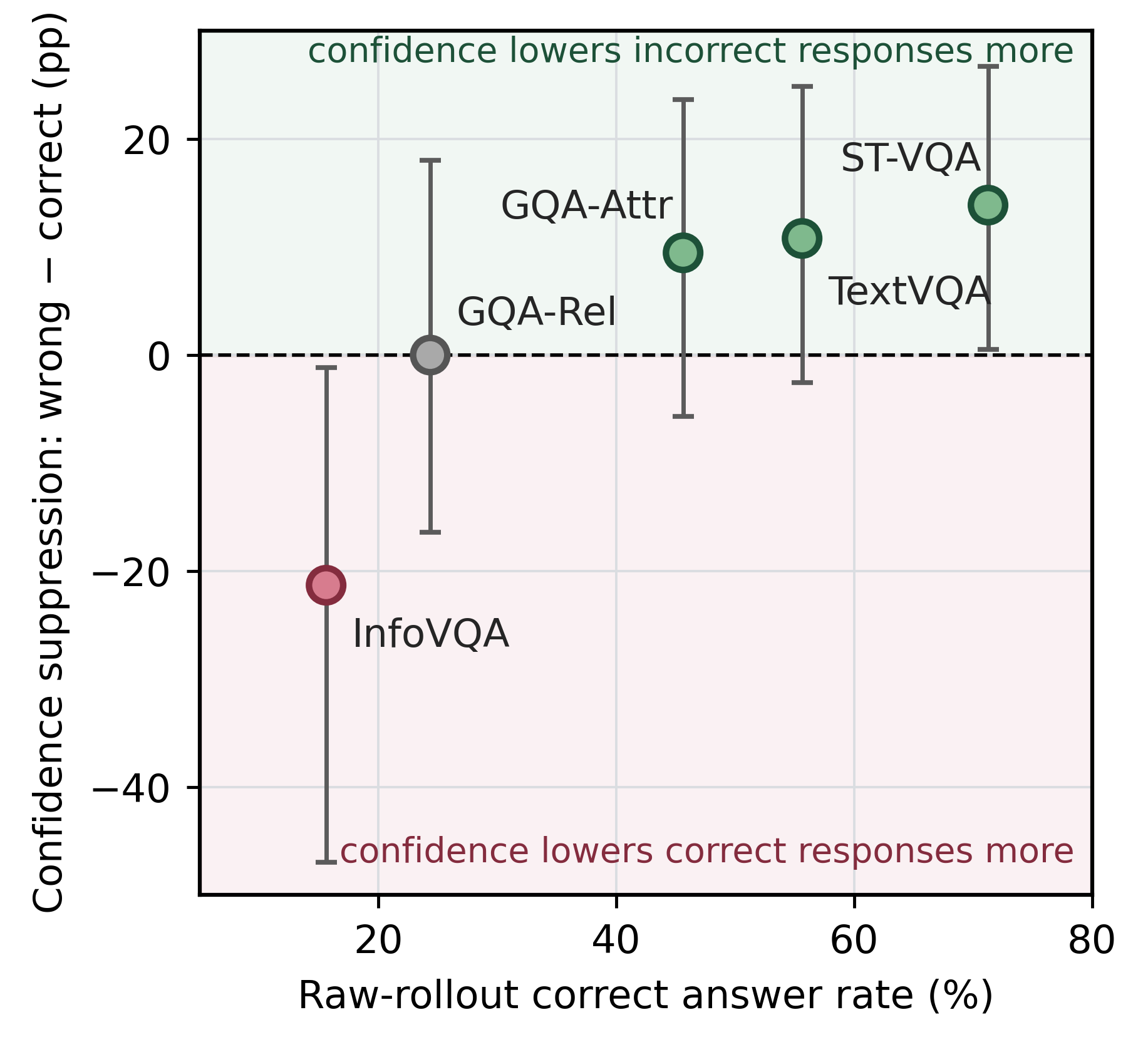}
    \caption{}
  \end{subfigure}

  \caption{Sensitivity to confidence strength and its limiting behavior. (a) Accuracy gain over TTRV with InternVL2.5-4B; hollow markers indicate the best strength for each task. (b) Relationship between the raw-rollout correct answer rate and the relative effect of confidence on incorrect and correct responses.}
  \label{fig:confidence_capacity}
\end{figure}

\paragraph{Confidence strength.}
To study sensitivity to the only method-specific hyperparameter, we vary the confidence strength $\lambda$ while keeping all other settings fixed. Figure~\ref{fig:confidence_capacity}(a) shows that \model{} outperforms TTRV throughout the sweep. Moderate values work best on TextVQA and GQA-Relation, whereas $\lambda=1$ reduces their gains; InfoVQA and
GQA-Attribute vary by less than $0.6$ points. Thus, TTIQ is generally robust to $\lambda$, but sensitive tasks benefit from a moderate rather than the strongest confidence constraint.

\paragraph{Confidence and initial task capacity.}
Figure~\ref{fig:confidence_capacity}(b) relates the raw-rollout correct-answer rate to the effect of full-strength confidence ($\lambda=1$). On InfoVQA, where only $15.6\%$ of rollouts are correct, confidence suppresses correct responses $21.3$ points more often than incorrect ones, potentially weakening the scarce useful training signals. In contrast, on ST-VQA, with a $71.2\%$ correct-answer rate, confidence suppresses incorrect responses $13.9$ points more often. This suggests that the role of confidence varies with the model's initial capacity on each task, motivating a controlled rather than uniformly strong confidence term.

\vspace{-7pt}
\section{Related Work}

\paragraph{LLM and VLM test-time reinforcement learning.}
Test-time training \cite{sun2020test,pmlr-v267-sun25h} adapts pretrained models using unlabeled target inputs and has recently been extended to large language-model (LLM) reasoning through reinforcement learning. Existing work in this direction has primarily focused on LLMs. TTRL uses majority-vote responses from repeated sampling as pseudo-labels \cite{zuo2026ttrl}; subsequent work improves this consensus signal using step-wise confidence \cite{wang2026beyond}, or analyzes how majority voting can suppress initially rare correct responses \cite{lin2026detecting}.
In comparison, test-time reinforcement learning for generative VLMs remains less explored. TTRV applies this paradigm to VLMs using response-frequency rewards and the empirical answer distribution \cite{singh2026ttrv}. Existing approaches therefore largely derive supervision from consistency within the model's own outputs \cite{he-etal-2025-llm}, leaving it unclear whether the resulting improvements reflect stronger multimodal understanding or merely greater output consistency.

\paragraph{Grounding in VLMs.}
Despite their strong performance, VLMs often produce responses without faithfully using the provided visual evidence. Recent analyses reveal different forms of this failure \cite{tong2024eyes,li2024naturalbench}. Mirage shows that frontier VLMs can retain substantial benchmark performance and generate detailed visual reasoning even without image access \cite{asadi2026mirage}. VLind-Bench further isolates language-prior-induced blindness through counterfactual tests \cite{lee2025vlind}. Motivated by these grounding failures, a complementary line of work leverages visual grounding signals to improve model predictions at inference time. VCD contrasts predictions under original and distorted images to reduce reliance on language priors \cite{leng2024mitigating}, while entropy-based methods use model uncertainty to identify or retrieve relevant visual evidence \cite{gropl2026entropy}. Grounding signals have also been incorporated into VLM training through fine-grained feedback and reward modeling \cite{yu2024rlhf,yan2024vigor}. Given both the prevalence of grounding failures and the demonstrated effectiveness of grounding signals in inference and training, their use as supervision for unlabeled test-time fine-tuning represents a promising yet comparatively underexplored direction.

\section{Conclusion}

We studied test-time reinforcement learning for VLMs and identified two limitations of output-consensus supervision. Motivated by these findings, we proposed \model{}, which estimates image and question dependence through controlled input ablations and uses the resulting signals for response-level reward construction and token-level credit assignment. Experiments across diverse VQA tasks and VLM scales show that \model{} achieves the best average performance at every evaluated scale. It also generalizes across VLM families and improves performance on unseen datasets without further training. These results suggest that image-question dependence learning signals provide a promising alternative to output consensus for VLM test-time reinforcement learning.

\bibliographystyle{plainnat}
\bibliography{main}
\appendix
%
\appendix
\raggedbottom
\makeatletter
\setlength{\@fptop}{0pt}
\setlength{\@dblfptop}{0pt}
\makeatother

\section{Dataset Statistics}
\label{app:dataset_statistics}

\paragraph{Evaluation splits.}
For each dataset, we use the official test split whenever its ground-truth annotations are publicly available. We randomly sample 20 examples from this split for test-time training and hide their answers throughout training; all remaining examples are used for evaluation. Although TTIQ does not access labels during training, ground-truth answers are still required to compute the evaluation metrics. Therefore, for datasets whose test annotations are not publicly released, we instead use the largest publicly available labeled validation or evaluation split and apply the same 20-example sampling protocol. Table~\ref{tab:appendix_dataset_splits} reports the exact source split and the resulting training and evaluation sizes for each dataset.

\begin{table*}[h]
  \centering
  
  \small
  \setlength{\tabcolsep}{5pt}
  \begin{tabular}{@{}l>{\raggedright\arraybackslash}p{0.24\textwidth}
    >{\raggedright\arraybackslash}p{0.25\textwidth}rr@{}}
    \toprule
    Dataset & Task focus & Public labeled source split
      & Unlabeled train & Evaluation \\
    \midrule
    InfoVQA      & Infographic reading and reasoning
      & Official test split & 20 & 3,268 \\
    TextVQA      & Scene-text question answering
      & Validation split & 20 & 4,980 \\
    GQA-Relation & Visual relations
      & Test-dev balanced, Relation subset & 20 & 3,873 \\
    GQA-Attribute& Visual attributes
      & Test-dev balanced, Attribute subset & 20 & 1,746 \\
    ST-VQA       & Scene-text question answering
      & Public train split & 20 & 26,054 \\
    MathVerse-VD & Vision-dominant mathematical reasoning
      & Test-mini, Vision-Dominant subset & 20 & 768 \\
    COCO-QA-N    & Counting
      & Test split, Number subset & 20 & 2,735 \\
    CLEVR-Humans & Compositional visual reasoning
      & Validation split & 20 & 7,182 \\
    \midrule
    Total        & & & 160 & 50,606 \\
    \bottomrule
  \end{tabular}
  \vspace{-5pt}
      \caption{Dataset statistics for the experiments. The 20 training answers are withheld from every test-time training method.}
      \vspace{-5pt}
      \label{tab:appendix_dataset_splits}
\end{table*}

\begin{figure}[!t]
  \centering

  \scriptsize
  \setlength{\tabcolsep}{2.8pt}
  \resizebox{\columnwidth}{!}{%
  \begin{tabular}{@{}clrrrrr@{}}
    \toprule
    Train \(n\) & Method & InfoVQA & TextVQA & GQA-Rel. & COCO-QA-N & Avg. \\
    \midrule
    20 & TTRV & 58.31 & 74.73 & 31.00 & 69.91 & 58.49 \\
    \rowcolor{blue!6}
       & \textbf{TTIQ} & \textbf{59.69} & \textbf{79.45}
       & \textbf{37.12} & \textbf{75.03} & \textbf{62.82} \\
    \midrule
    50 & TTRV & \textbf{59.03} & 77.12 & 35.43 & 73.56 & 61.29 \\
    \rowcolor{blue!6}
       & \textbf{TTIQ} & 58.85 & \textbf{79.14}
       & \textbf{37.33} & \textbf{73.71} & \textbf{62.26} \\
    \midrule
    100 & TTRV & 57.94 & 77.86 & \textbf{38.20} & 72.81 & 61.70 \\
    \rowcolor{blue!6}
        & \textbf{TTIQ} & \textbf{58.47} & \textbf{80.43}
        & 37.36 & \textbf{74.16} & \textbf{62.60} \\
    \bottomrule
  \end{tabular}
  }
    \captionof{table}{Exact match accuracy (\%) with 20, 50, and 100 unlabeled training examples. Bold marks the better method for each training size and task.}
    \label{tab:appendix_support_size}
\end{figure}

\section{Training-Set Size and Efficiency}
\label{app:support_and_efficiency}

\subsection{Impact of Training-Set Size}
\label{app:support_size}

We evaluate TTIQ and TTRV using nested sets of 20, 50, and 100 unlabeled examples on four representative tasks with InternVL2.5-4B. All settings use the same held-out split obtained by excluding the full 100-example training pool. We increase the number of updates proportionally with the training-set size, keeping the average number of exposures per example approximately constant.

\begin{figure}[!t]
  \centering
  \begin{subfigure}[t]{0.48\linewidth}
    \centering
    \includegraphics[width=\linewidth]
      {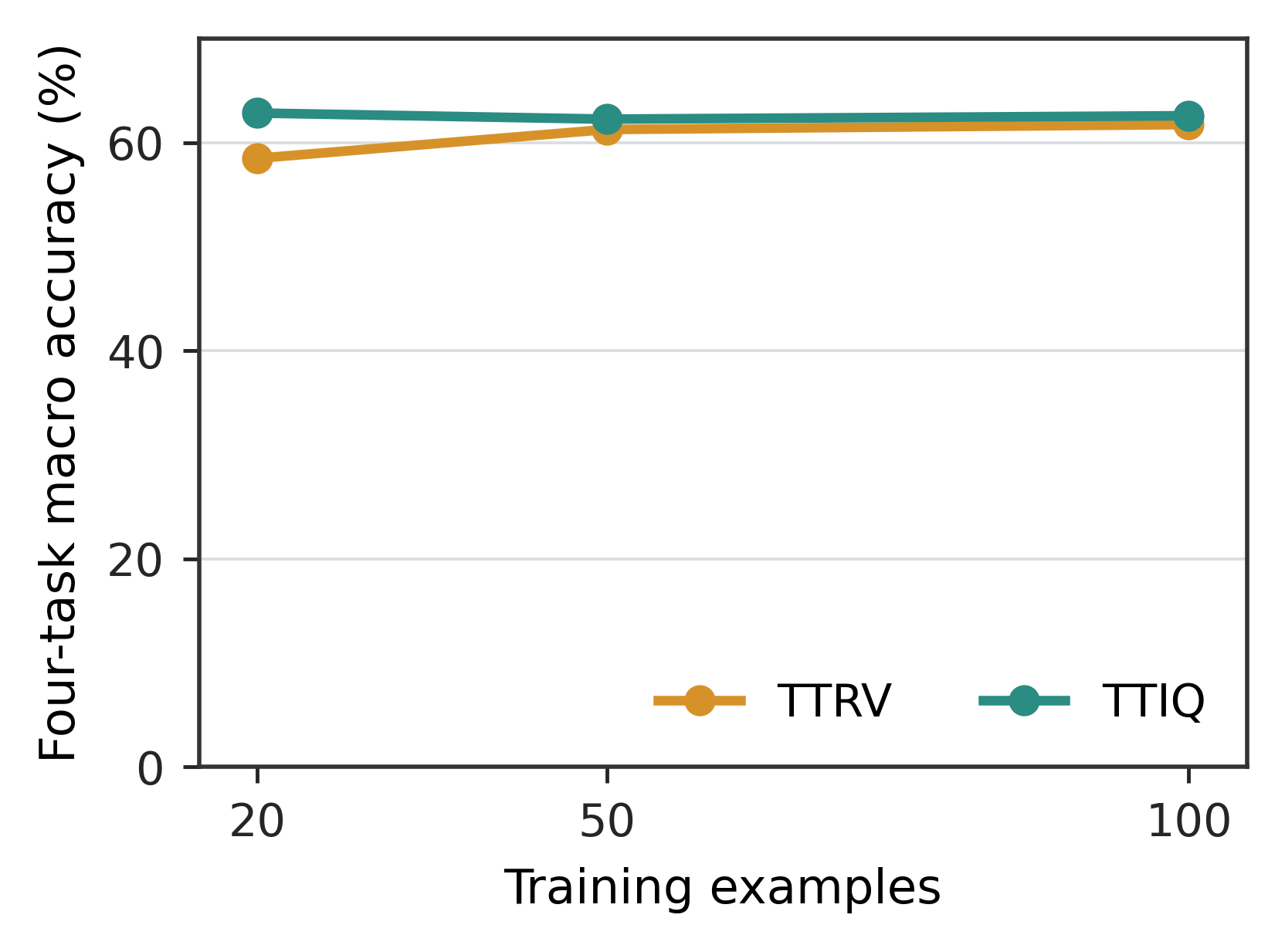}
    \caption{}
  \end{subfigure}
  \hfill
  \begin{subfigure}[t]{0.48\linewidth}
    \centering
    \includegraphics[width=\linewidth]
      {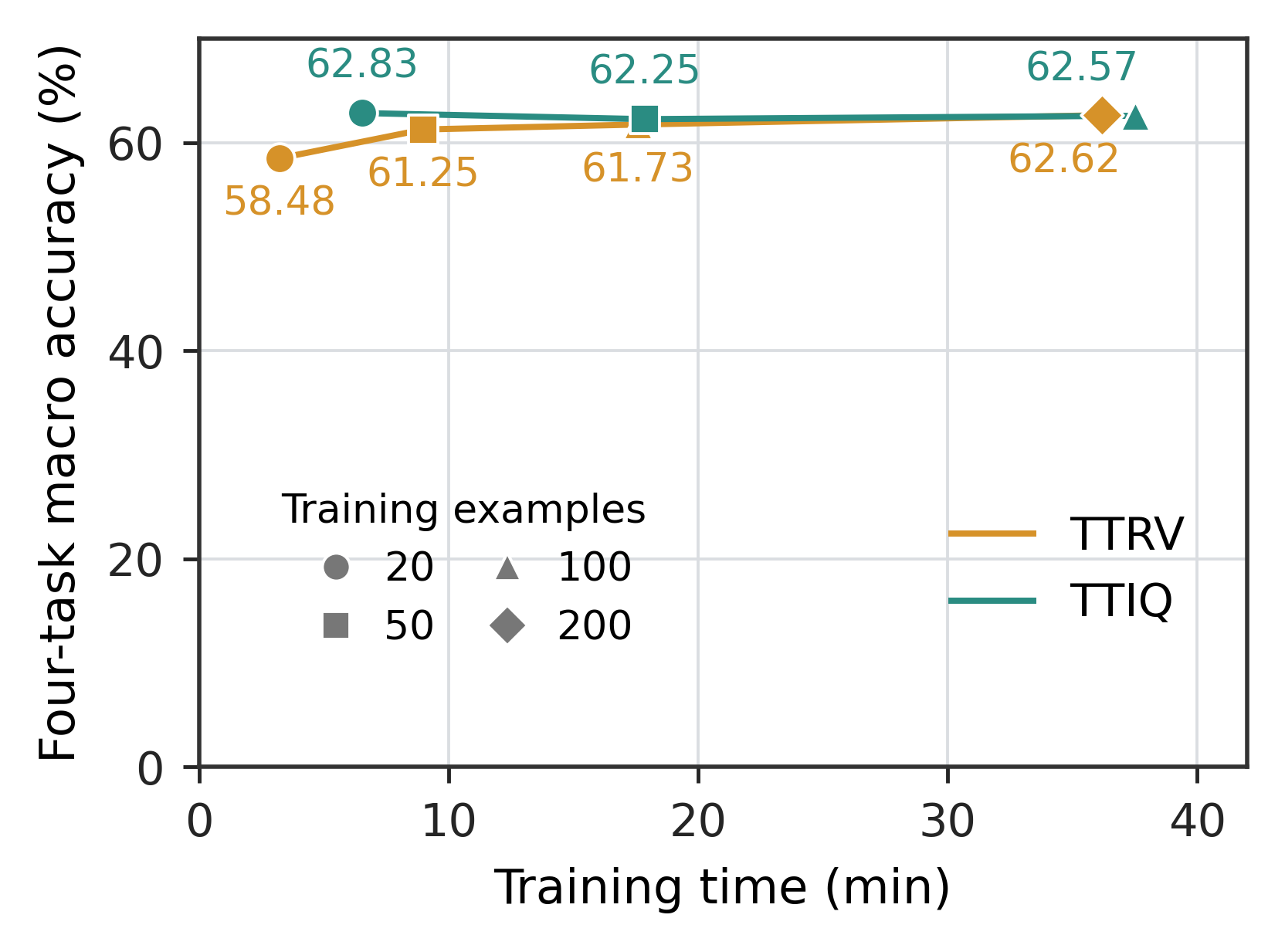}
    \caption{}
  \end{subfigure}
  \caption{Effect of training-set size with InternVL2.5-4B.
  \textbf{(a)} Macro accuracy over InfoVQA, TextVQA, GQA-Relation, and
  COCO-QA-N.
  \textbf{(b)} Mean training time on four GH200 GPUs.
  Training time includes only the update loop; model setup and evaluation are
  excluded.}
  \label{fig:appendix_support_size_tradeoff}
\end{figure}

Table~\ref{tab:appendix_support_size} and
Figure~\ref{fig:appendix_support_size_tradeoff} show distinct scaling behavior. TTIQ achieves strong macro result with only 20 examples and remains broadly stable as more examples are added. In contrast, TTRV improves more with training-set size, but its result with 100 examples still remains below that of TTIQ with 20 examples. TTIQ also wins on most individual task--size combinations. This contrast reflects a fundamental difference in their supervision signals. TTRV relies on response agreement as an indirect proxy for correctness, whose useful effects may accumulate over more training examples. In contrast, TTIQ uses controlled input ablations to expose, within each example, whether a response is grounded in both the image and the question. It therefore directly corrects how the model uses multimodal evidence rather than relying on a useful training signal to emerge across many examples. Once this effective signal is available, increasing the training set mainly broadens example coverage, resulting in limited and task-dependent gains beyond 20 examples.

\subsection{Training Efficiency}
\label{app:training_efficiency}

Table~\ref{tab:appendix_training_time} reports a matched training-only comparison on InfoVQA. The two additional teacher-forced scoring passes increase the update time, but do not increase peak GPU memory. With eight updates, TTIQ finishes in 4.78 minutes, compared with 2.92 minutes for TTRV, while both use the same peak memory.

Figure~\ref{fig:appendix_support_size_tradeoff} further shows that training time grows approximately linearly with the number of updates. Because test-time training is performed on a small unlabeled subset, the absolute training cost remains modest, making the additional scoring overhead of TTIQ acceptable in practice. In particular, the 20-example setting provides a favorable accuracy--efficiency trade-off.

\begin{figure}[!t]
  \centering

  \small
  \setlength{\tabcolsep}{3pt}
  \begin{tabular}{@{}lrrrr@{}}
    \toprule
    Method & Extra TF & Sec./update & 8 updates & Peak GB/GPU \\
    \midrule
    TTRV & 0 & 21.9 & 2.92 min & 76.84 \\
    \rowcolor{blue!6}
    \textbf{TTIQ} & 2 & 35.8 & 4.78 min & 76.84 \\
    \bottomrule
  \end{tabular}
   \captionof{table}{Matched training-only efficiency on InternVL2.5-4B and InfoVQA with four GH200 GPUs.}
    \label{tab:appendix_training_time}
\end{figure}

\begin{center}
  \centering
  \captionsetup{type=table}

  \footnotesize
  \setlength{\tabcolsep}{4pt}
  \begin{tabular}{@{}llrr@{}}
      \toprule
      Backbone & Method & Token-F1 & ROUGE-L \\
      \midrule
      InternVL3-2B & Base  & 61.65 & 61.56 \\
                   & TTRL  & 62.65 & 62.58 \\
                   & SCOPE & 62.66 & 62.59 \\
                   & TTRV  & 62.16 & 62.07 \\
      \rowcolor{blue!6}
                   & \textbf{TTIQ} & \textbf{63.53} & \textbf{63.45} \\
      \midrule
      InternVL2.5-4B & Base  & 59.94 & 59.87 \\
                     & TTRL  & 65.41 & 65.36 \\
                     & SCOPE & 65.55 & 65.49 \\
                     & TTRV  & 65.23 & 65.17 \\
      \rowcolor{blue!6}
                     & \textbf{TTIQ} & \textbf{66.96} & \textbf{66.89} \\
      \midrule
      InternVL3-8B & Base  & 70.22 & 70.14 \\
                   & TTRL  & 69.88 & 69.82 \\
                   & SCOPE & 69.75 & 69.69 \\
                   & TTRV  & 69.79 & 69.73 \\
      \rowcolor{blue!6}
                   & \textbf{TTIQ} & \textbf{70.76} & \textbf{70.69} \\
    \bottomrule
  \end{tabular}
    \caption{Macro Token-F1 and ROUGE-L F1 over the eight complete
  evaluation sets. Best results within each backbone are bold.}
  \label{tab:appendix_soft_metrics}
\end{center}

\section{Beyond Exact Match}
\label{app:partial_credit}
\begin{table*}[!t]
  \centering
  \small

  \small
  \setlength{\tabcolsep}{4.5pt}
  \begin{tabularx}{\textwidth}
    {@{}lX>{\raggedright\arraybackslash}p{0.25\textwidth}lll@{}}
    \toprule
    Task & Question & Base & TTRV & TTIQ & GT \\
    \midrule
    GQA-Rel.
      & What vehicle is in front of the red vehicle?
      & A car. & Car. & \textbf{SUV} & suv \\
    GQA-Rel.
      & What are the jars sitting on top of?
      & Tea cups. & Cups. & \textbf{Stove.} & stove \\
    GQA-Attr
      & How clean are the pots?
      & The pots are clean. & Clean. & \textbf{Dirty.} & dirty \\
    GQA-Attr
      & Which side of the image is the cyclist on?
      & Left. & Left & \textbf{Right.} & right \\
    ST-VQA
      & What number is this bus known as?
      & Private bus. & Private. & \textbf{23} & 23 \\
    ST-VQA
      & What does the advertisement say behind the woman's head?
      & Stretchy shery & Stretchy & \textbf{Stretch} & Stretch \\
    COCO-QA-N
      & How many giraffes are standing near each other in an enclosed area?
      & Three giraffes are standing near each other in an enclosed area.
      & 3 & \textbf{4} & four \\
    COCO-QA-N
      & How many people are on snow skis while one also has wheels?
      & Two people on snow skis, one of whom has a wheelchair.
      & 2 & \textbf{3} & three \\
    CLEVR-H
      & What color is the metallic cylinder?
      & Blue. & The metallic cylinder is blue. & \textbf{Red} & red \\
    \bottomrule
  \end{tabularx}
    \caption{Representative form-to-content correction chains from the final
  InternVL3-2B outputs. TTRV changes the surface form or length of the Base
  response but retains the same central incorrect answer, whereas TTIQ
  changes the answer content to an acceptable ground truth.}
  \label{tab:appendix_semantic_corrections}
\end{table*}

\definecolor{TTIQWrong}{HTML}{A83232}
\definecolor{TTIQCorrect}{HTML}{247A45}

\newcommand{\TTIQAppendixCase}[8]{%
  \begin{minipage}[t]{0.49\textwidth}
    \begin{minipage}[t]{0.30\linewidth}
      \vspace{0pt}\centering #1
    \end{minipage}\hfill
    \begin{minipage}[t]{0.67\linewidth}
      \vspace{0pt}
      {\small\textbf{#2}}\\[-0.1em]
      {\scriptsize\textbf{Q:} #3}\\[0.12em]
      {\scriptsize\color{TTIQWrong}\textbf{\(\times\) #4:} ``#5''}\\[-0.08em]
      {\scriptsize\color{TTIQWrong}\hspace*{1em}#6}\\[-0.02em]
      {\scriptsize\color{TTIQCorrect}\textbf{\(\checkmark\) Joint:}
        ``#7''}\\[-0.08em]
      {\scriptsize\color{TTIQCorrect}\hspace*{1em}#8}
    \end{minipage}
  \end{minipage}%
}

\begin{figure*}[!t]
  \centering
  \TTIQAppendixCase
    {\includegraphics[width=\linewidth,height=2.15cm,keepaspectratio,
      trim=520 145 220 180,clip]
      {Fig/appendix/intro_question_misaligned_infovqa_4b.jpeg}}
    {Image salience without question alignment (InfoVQA)}
    {When was the second-generation 8080 chip produced?}
    {Image-only}{1972}{\(I=+1.00,\ Q=-1.00\)}
    {1974}{\(I=-.33,\ Q=+.33\)}
  \hfill
  \TTIQAppendixCase
    {\includegraphics[width=\linewidth,height=2.15cm,keepaspectratio]
      {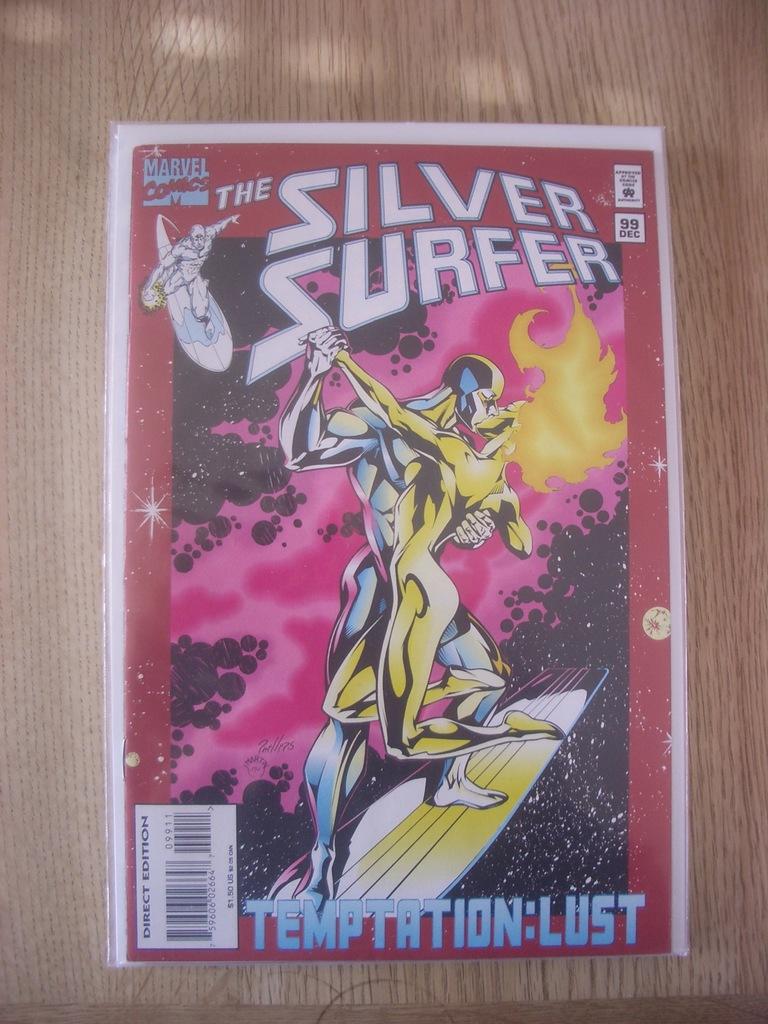}}
    {Image salience without question alignment (TextVQA)}
    {What is the name of this issue?}
    {Image-only}{99 DEC}{\(I=+1.00,\ Q=-1.00\)}
    {Temptation: Lust}{\(I=.00,\ Q=.00\)}

  \par\vspace{0.65em}\noindent
  \TTIQAppendixCase
    {\includegraphics[width=\linewidth,height=2.15cm,keepaspectratio,
      trim=50 170 60 0,clip]
      {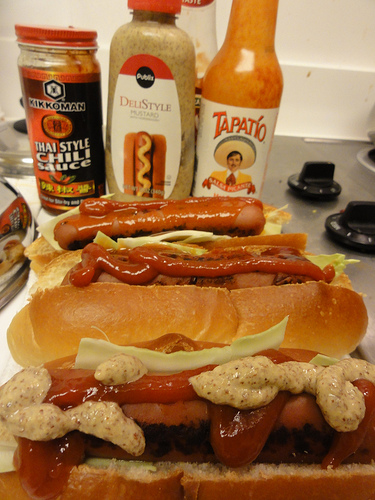}}
    {Image salience without question alignment (ST-VQA)}
    {What is the brand of mustard?}
    {Image-only}{DeliStyle Spanish}{\(I=+1.00,\ Q=-1.00\)}
    {Publix}{\(I=-.33,\ Q=+.33\)}
  \hfill
  \TTIQAppendixCase
    {\includegraphics[width=\linewidth,height=2.15cm,keepaspectratio]
      {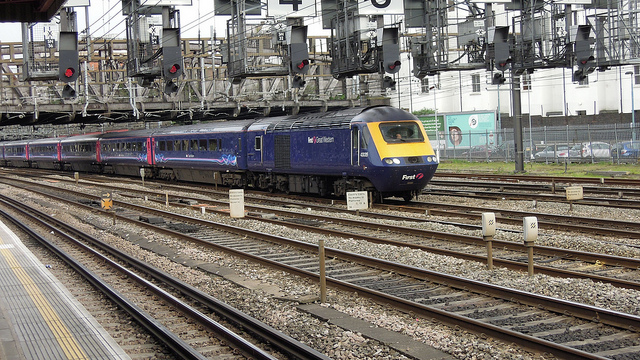}}
    {Image salience without question alignment (COCO-QA)}
    {How many trains are on the multiple tracks side by side?}
    {Image-only}{5}{\(I=+.33,\ Q=-1.00\)}
    {1}{\(I=+1.00,\ Q=+1.00\)}

  \par\vspace{0.65em}\noindent
  \TTIQAppendixCase
    {\includegraphics[width=\linewidth]
      {Fig/appendix/appendix_last_player_textvqa.jpg}}
    {Weak image dependence (TextVQA)}
    {What is the number on the last player in blue?}
    {Question-only}{16}{\(I=-1.00,\ Q=-1.00\)}
    {19}{\(I=+.67,\ Q=+.67\)}
  \hfill
  \TTIQAppendixCase
    {\includegraphics[width=\linewidth]
      {Fig/appendix/appendix_blanket_gqa_72dpi.jpg}}
    {Weak image dependence (GQA)}
    {What is the color of the soft blanket?}
    {Question-only}{green}{\(I=-1.00,\ Q=+.33\)}
    {white}{\(I=+1.00,\ Q=+1.00\)}

  \par\vspace{0.65em}\noindent
  \TTIQAppendixCase
    {\includegraphics[width=\linewidth,height=2.15cm,keepaspectratio]
      {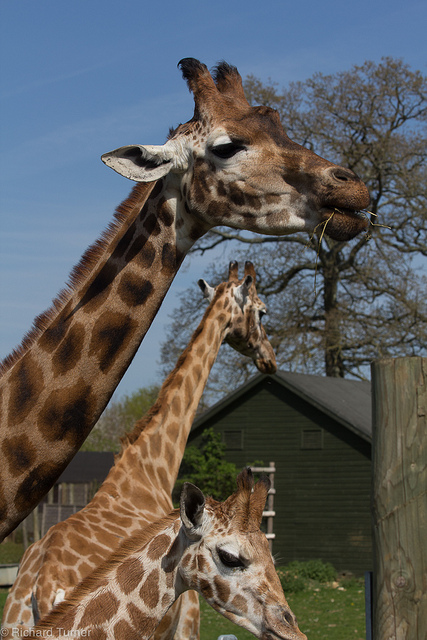}}
    {Weak image dependence (COCO-QA)}
    {How many large giraffes are standing next to the barn?}
    {Question-only}{two}{\(I=-.33,\ Q=-.33\)}
    {three}{\(I=+.67,\ Q=+.67\)}
  \hfill
  \TTIQAppendixCase
    {\includegraphics[width=\linewidth,height=2.15cm,keepaspectratio]
      {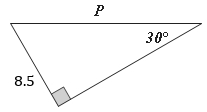}}
    {Weak image dependence (MathVerse)}
    {Solve for the unknown \(P\) in the adjacent figure.}
    {Question-only}{10}{\(I=-1.00,\ Q=+.33\)}
    {17}{\(I=+1.00,\ Q=+1.00\)}

\caption{
Qualitative examples of image--question dependence before test-time training. Each example compares two responses sampled for the same image--question input. Incorrect responses either follow visual content without sufficient question alignment (top) or exhibit weak image dependence (bottom), while the correct responses depend more strongly on both inputs.
}
  \label{fig:appendix_dependence_examples}
\end{figure*}

\let\TTIQAppendixCase\relax

\subsection{Additional Evaluation Metrics}
\label{app:soft_metrics}

We further report Token-F1 and ROUGE-L F1 alongside exact-match accuracy. These metrics provide a more graded evaluation by measuring token and sequence overlap between the prediction and the ground-truth answer. Let \(N(z)\) denote the token sequence obtained after applying the same normalization as the exact-match evaluator.

For a prediction \(\hat{y}\) and a ground-truth answer \(a\), we define their multiset token overlap as
\begin{equation}
O(\hat{y},a)
=
\sum_{w}
\min\!\left(
c_w\!\left(N(\hat{y})\right),
c_w\!\left(N(a)\right)
\right),
\end{equation}
where \(c_w(\cdot)\) denotes the number of occurrences of token \(w\). Token-level precision, recall, and F1 are then defined as
\begin{equation}
\begin{aligned}
P_{\mathrm{tok}}
&=
\frac{O(\hat{y},a)}{|N(\hat{y})|},
\qquad
R_{\mathrm{tok}}
=
\frac{O(\hat{y},a)}{|N(a)|},
\\
F_{\mathrm{tok}}
&=
\frac{2P_{\mathrm{tok}}R_{\mathrm{tok}}}
     {P_{\mathrm{tok}}+R_{\mathrm{tok}}}.
\end{aligned}
\end{equation}

For ROUGE-L, let
\begin{equation}
L(\hat{y},a)
=
\operatorname{LCS}\!\left(N(\hat{y}),N(a)\right)
\end{equation}
denote the length of the longest common subsequence between the normalized prediction and ground-truth answer. ROUGE-L precision, recall, and F1 are defined as
\begin{equation}
\begin{aligned}
P_{\mathrm{L}}
&=
\frac{L(\hat{y},a)}{|N(\hat{y})|},
\qquad
R_{\mathrm{L}}
=
\frac{L(\hat{y},a)}{|N(a)|},
\\
F_{\mathrm{L}}
&=
\frac{2P_{\mathrm{L}}R_{\mathrm{L}}}
     {P_{\mathrm{L}}+R_{\mathrm{L}}}.
\end{aligned}
\end{equation}

When an example contains multiple ground-truth answers, we take the maximum score over the ground-truth answer set \(\mathcal{G}\):
\begin{equation}
\begin{aligned}
S_{\mathrm{tok}}(\hat{y},\mathcal{G})
&=
\max_{a\in\mathcal{G}} F_{\mathrm{tok}}(\hat{y},a),
\\
S_{\mathrm{L}}(\hat{y},\mathcal{G})
&=
\max_{a\in\mathcal{G}} F_{\mathrm{L}}(\hat{y},a).
\end{aligned}
\end{equation}
\begin{center}
  \centering
  \captionsetup{type=figure}
  \includegraphics[width=0.94\linewidth]
    {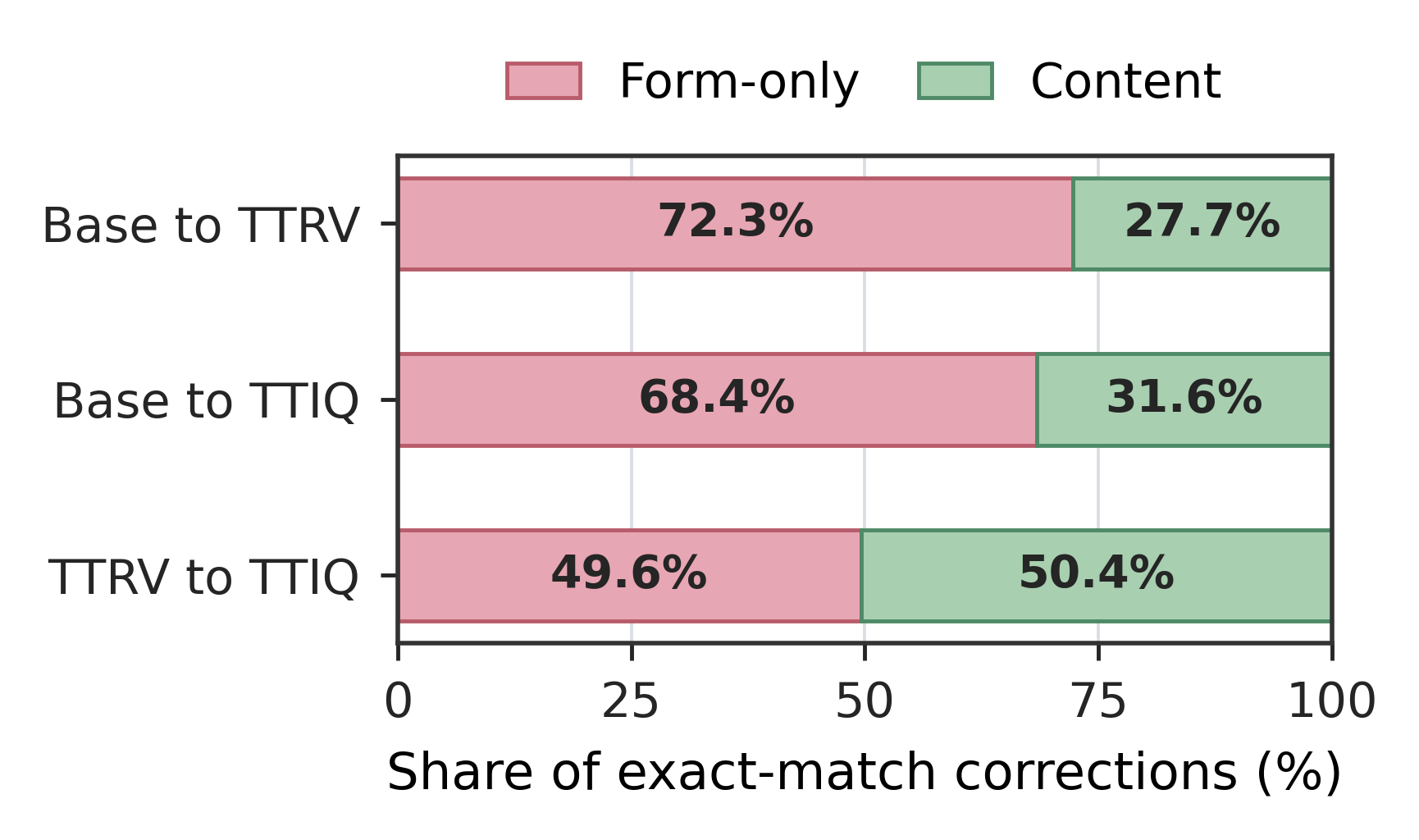}
  \caption{
Composition of exact-match corrections. ``Base to TTRV'' denotes examples answered incorrectly by the base model but correctly by TTRV; ``Base to TTIQ'' is defined analogously. ``TTRV to TTIQ'' denotes examples for which TTRV remains incorrect but TTIQ becomes correct. Each bar shows the proportion of form-only and content-changing corrections within the corresponding correction set.
}
  \label{fig:appendix_correction_composition}
\end{center}

Table~\ref{tab:appendix_soft_metrics} reports macro averages over the eight complete evaluation sets. TTIQ achieves the highest average Token-F1 and ROUGE-L F1 for every backbone. Since these metrics place greater emphasis on content overlap and are less sensitive to minor differences in answer form, the results suggest that TTIQ's gains extend beyond answer-form changes to better alignment with the ground-truth content.

\subsection{Correction Composition}
\label{app:correction_types}

We further analyze the exact-match corrections using the diagnostic defined in Section~3.1, which distinguishes form-only corrections from corrections that change the predicted content. As shown in
Figure~\ref{fig:appendix_correction_composition}, both TTRV and TTIQ correct many form-related errors, but TTIQ produces a larger proportion of content corrections relative to the base model. More importantly, among cases where TTRV remains incorrect but TTIQ becomes correct, content corrections account for slightly more than half of the improvement. Thus, TTIQ preserves the benefit of improved answer formulation while extending the gains to more substantive changes in the predicted answer, consistent with its stronger Token-F1 and ROUGE-L results.

\section{Additional Qualitative Examples}
\label{app:qualitative_examples}

\subsection{Additional Content Corrections}
\label{app:qualitative_corrections}

Table~\ref{tab:appendix_semantic_corrections} supplements the first analysis in Section~3.1 with cases not shown in the main paper. In every row, Base and TTRV retain the same incorrect content, whereas TTIQ changes the predicted object, attribute, count, text, or mathematical decision to an acceptable answer. The table covers all eight evaluation tasks and seven further cases.

\subsection{Examples of Joint Image--Question Dependence}
\label{app:qualitative_dependence}

Figure~\ref{fig:appendix_dependence_examples} presents eight additional image--question examples, each with two responses sampled from InternVL2.5-4B before test-time training. The displayed \(I\) and \(Q\) values measure how strongly each response depends on the image and the question, respectively.

In the top four examples, the incorrect response has strong image dependence but weak question dependence. It follows salient visual content without correctly addressing the question. In the bottom four examples, the incorrect response has weak image dependence and relies more heavily on textual cues or language priors, resulting in an answer that is not sufficiently supported by the image. These examples show that strong dependence on either input alone is insufficient. TTIQ therefore rewards responses only when they depend strongly on both the image and the question. Across all examples, the response with stronger joint dependence gives the correct answer.


\end{document}